%% file: main.tex
\documentclass[10pt,twocolumn,letterpaper]{article}

\usepackage[pagenumbers]{cvpr} 

\usepackage{graphicx}
\usepackage{amsmath}
\usepackage{amssymb}
\usepackage{booktabs}
\usepackage{multirow}

\usepackage[pagebackref,breaklinks,colorlinks]{hyperref}

\usepackage[capitalize]{cleveref}
\crefname{section}{Sec.}{Secs.}
\Crefname{section}{Section}{Sections}
\Crefname{table}{Table}{Tables}
\crefname{table}{Tab.}{Tabs.}

\def\cvprPaperID{11635} 
\def\confName{CVPR}
\def\confYear{2022}

\input{preamble}
\begin{document}
\input{sec/0_metadata}
\maketitle
\input{sec/0_abstract}
\input{sec/1_introduction}

\input{sec/2_model}

\input{sec/3_data}

\input{sec/4_results}

\input{sec/5_conclusions}

{
    \clearpage
    \small
    \bibliographystyle{ieee_fullname}
    \bibliography{macros,main}
}

\input{sec/X_supplementary}


\end{document}

%% file: preamble.tex
\usepackage{overpic}
\usepackage{enumitem} 
\usepackage{overpic} 
\usepackage{color}

\definecolor{turquoise}{cmyk}{0.65,0,0.1,0.3}
\definecolor{purple}{rgb}{0.65,0,0.65}
\definecolor{dark_green}{rgb}{0, 0.5, 0}
\definecolor{orange}{rgb}{0.8, 0.6, 0.2}
\definecolor{red}{rgb}{0.8, 0.2, 0.2}
\definecolor{darkred}{rgb}{0.6, 0.1, 0.05}
\definecolor{blueish}{rgb}{0.0, 0.3, .6}
\definecolor{light_gray}{rgb}{0.7, 0.7, .7}
\definecolor{pink}{rgb}{1, 0, 1}
\definecolor{greyblue}{rgb}{0.25, 0.25, 1}

\usepackage{blindtext}

\renewcommand{\paragraph}[1]{\vspace{1em}\noindent\textbf{#1}.}

\newcommand{\PAR}[1]{\vskip4pt \noindent{\bf #1~}}

%% file: sec/0_metadata.tex
\title{DeepAD: A Robust Deep Learning Model of Alzheimer's Disease \\ Progression for Real-World Clinical Applications}

\author{Somaye Hashemifar,
        Claudia Iriondo, Mohsen Hejrati;\\   
        for the Alzheimer’s Disease Neuroimaging Initiative\thanks{Data used in preparation of this article were obtained from the Alzheimer's Disease Neuroimaging Initiative (ADNI) database (\href{http://adni.loni.usc.edu}{adni.loni.usc.edu}). As such, the investigators within the ADNI contributed to the design and implementation of ADNI and/or provided data but did not participate in analysis or writing of this report. A complete listing of ADNI investigators can be found at \url{http://adni.loni.usc.edu/wpcontent/uploads/how_to_apply/ADNI_Acknowledgement_List.pdf} } \\
        {\tt\small \{hashems4, iriondoc, hejratis\}@gene.com}
        }


%% file: sec/0_abstract.tex
\begin{abstract}
The ability to predict the future trajectory of a patient is a key step toward the development of therapeutics for complex diseases such as Alzheimer’s disease (AD). However, most machine learning approaches developed for prediction of disease progression are either single-task or single-modality models, which can not be directly adopted to our setting involving multi-task learning with high dimensional images. Moreover, most of those approaches are trained on a single dataset (i.e. cohort), which can not be generalized to other cohorts. We propose a novel multimodal multi-task deep learning model to predict AD progression by analyzing longitudinal clinical and neuroimaging data from multiple cohorts. Our proposed model integrates high dimensional MRI features from a 3D convolutional neural network  with other data modalities, including clinical and demographic information, to predict the future trajectory of patients. Our model employs an adversarial loss to alleviate the study-specific imaging bias, in particular the inter-study domain shifts.  In addition, a Sharpness-Aware Minimization (SAM) optimization technique is applied to further improve model generalization. The proposed model is trained and tested on various datasets in order to evaluate and validate the results. Our results showed that 1) our model yields significant improvement over the baseline models, and 2) models using extracted neuroimaging features from 3D convolutional neural network outperform the same models when applied to MRI-derived volumetric features. 
\end{abstract}

%% file: sec/1_introduction.tex
\section{Introduction}
\label{sec:intro}

\textbf{Alzheimer's disease (AD)} is the most common cause of dementia in people over 65, with 26.6 million people suffering worldwide \cite{brookmeyer2007forecasting}. AD is a slowly progressing disease caused by the degeneration of brain cells, with patients showing clinical symptoms years after the onset of the disease. Therefore, accurate diagnosis and treatment of AD in its early stage, i.e., mild cognitive impairment (MCI), is critical to prevent non-reversible and fatal brain damage.

\input{fig/ad-stages}

Existing diagnostic methods have focused on the task of classifying patients into coarse categories:  Cognitive Normal (CN), Mild Cognitive Impairment (MCI), or Alzheimer's Disease (AD). However, real-word applications require a more fine-grained measurement scale. One way to solve this is to instead predict the outcome of cognitive and functional tests such as Clinical Dementia Rating Scale Sum of Boxes (CDRSB), Alzheimer’s Disease Assessment Scale-Cognitive Subscale (ADAS-COG12), and Mini-Mental State Examination (MMSE) which are measured by continuous numerical values. This approach helps to provide more granular estimations of disease progression but can be very noisy and subjective across studies -- making it challenging to establish treatment efficacy in small patient populations.

The heterogeneous nature of AD is also an important consideration when predicting progression. An accurate model of AD progression must take into account multiple modalities such as environmental factors, genomics, demographics, and brain imaging \cite{alberdi2016early} and simultaneously model the complex interactions between each modality. Patients with AD may present different phenotypes and progression patterns, requiring different therapies. 

In this paper we focus on the task of modeling the trajectory of AD over time as measured by three cognitive tests (CDRSB, ADAS-COG12, MMSE) using a multimodal approach that uses cognitive scores, genomic, demographic, and imaging data as input. This is a fundamental building block toward developing and testing new therapeutics that require fine-grained estimation of treatment effects. With an accurate multimodal forward model of AD progression, we can gain insight into the interactions between each of the observed features -- enabling more treatments to be developed at a lower cost.

Despite demand, little progress has been made because of the difficult design requirements and lack of large-scale, homogeneous datasets that contain early stage AD patients. Much of the prior work has focused on using image-derived features (using FreeSurfer, ANTS) in order to overcome the high amount of variability in raw MRI images and small dataset sizes. 

In this study we propose \textbf{DeepAD}, a multimodal multi-task deep neural network for personalized prediction of disease progression and diagnosis. Our model leverages longitudinal imaging and clinical data to generate a patient-specific trajectory of progression. These predictions can be used as a way to make an accurate prognosis for reflecting each individual patient’s dynamic disease state.  Comprehensive evaluations were conducted on varying datasets to ensure our findings generalize across several domains i.e., different source datasets and patient disease states. The major contributions of our work are as follows:

\PAR{\textbf{Multimodality:}} Our model combines various inputs including cognitive scores, demographics, genomics, and 3D MR images, which enables learning of complimentary information across modalities for a large patient population. Tabular data and imaging data are fed as separate inputs to the model, the feature representations are fused downstream, and training is performed end to end to construct a feature space for predicting patients' progression. 

\PAR{\textbf{Robustness to domain shift (Generalization):}}
\begin{itemize}
\item \textbf{Adversarial loss:} Convolutional neural networks trained on multi-study MR imaging data often suffer from problems arising from domain shift and heterogeneity, as scanners and  protocols may vary by study site and the patient population and their disease state may differ between studies. We trained a 3D convolutional neural network using adversarial loss as proposed in \cite{kim2019learning, ganin2015unsupervised} to extract neuroimaging features from full-resolution (3D) MR images while remaining invariant to study site domain.

\item \textbf{Sharpness-aware minimization (SAM):} for overparameterized deep neural networks, optimization methods such as SGD and ADAM can converge to local minima, which lead to poor test time performance \cite{foret2020sharpness}.  SAM \cite{foret2020sharpness} is incorporated into our base optimization method, either ADAM or SGD, to avoid overfitting and improve generalization. SAM performs two forward-backward passes to estimate the smoothness of the loss landscape and improve final prediction accuracy.

\end{itemize}

\PAR{\textbf{Integrating Longitudinal data:}} Our model learns to predict change in cognitive test scores (CDRSB, ADAS-COG12, MMSE) one year in the future from the beginning of the study, a timeframe that is particularly relevant for clinical trials aiming to halt AD progression. However, since this severely limits the amount of data that can be used for training, we also make use of a pretraining step that operates on all timepoints. Integrating the longitudinal features allows for learning the underlying temporal characteristics in the disease. These features enable  our model to learn a better understanding of the individual disease progression that contributes to more accurate diagnosis, monitoring, and prognosis, and thus is beneficial for and applicable to personalized medicine. 


\section{Related work}

Over the past decade, machine learning and deep learning based approaches, including the support vector machine, random forest, recurrent neural network (RNN), and convolutional neural networks (CNN) have been proposed for prognosis, predicting disease progression, monitoring treatment effects and for stratifying AD patients \cite{moore2019random, moradi2015machine, zhang2012predicting, zhang2012multi}. A multi-modal GRU-based RNN was used to integrate longitudinal clinical information and cross-sectional tabular imaging features for classifying the MCI patients into converter to AD or not-converter to AD \cite{lee2019predicting}. MinimalRNN \cite{chen2017minimalrnn} employed similar features for regressing a couple of endpoints and for stratifying patients to cognitively normal (CN), MCI, and AD \cite{nguyen2020predicting}. An ensemble model based on stacked CNN and a bidirectional long short-term memory (BiLSTM) was utilized to jointly predict multiple endpoints on the fusion of time series clinical features and Freesurfer derived imaging features \cite{el2020multimodal}. Recently, several methods have started to employ 3D CNN based models to extract features from MRIs. A 3D convolutional autoencoder (CAE) with transfer learning is employed on MRIs to stratify the patients into progressor and non-progressor groups \cite{oh2019classification}. A stacked denoising auto-encoder approach was used to extract features from clinical and genetic data, and a 3D CNN for MRIs to categorize patients into different stages of the disease \cite{venugopalan2021multimodal}. 

Previous research has focused on developing single-task and/or single modality models such as predicting MCI to AD conversion \cite{lee2019predicting}, classification into AD stages \cite{venugopalan2021multimodal, qiu2018fusion} , or predicting one or few cognitive endpoints \cite{nguyen2020predicting, el2020multimodal}, which is not applicable to personalized medicine for AD. Besides, single-task and single-modalities models exploit neither the complementary information among modalities nor the correlation between tasks. Developing models to overcome such limitations is fundamental to advancing personalized medicine.

%% file: fig/ad-stages.tex
\begin{figure*}
\begin{center}
\includegraphics[width=1.99\columnwidth]{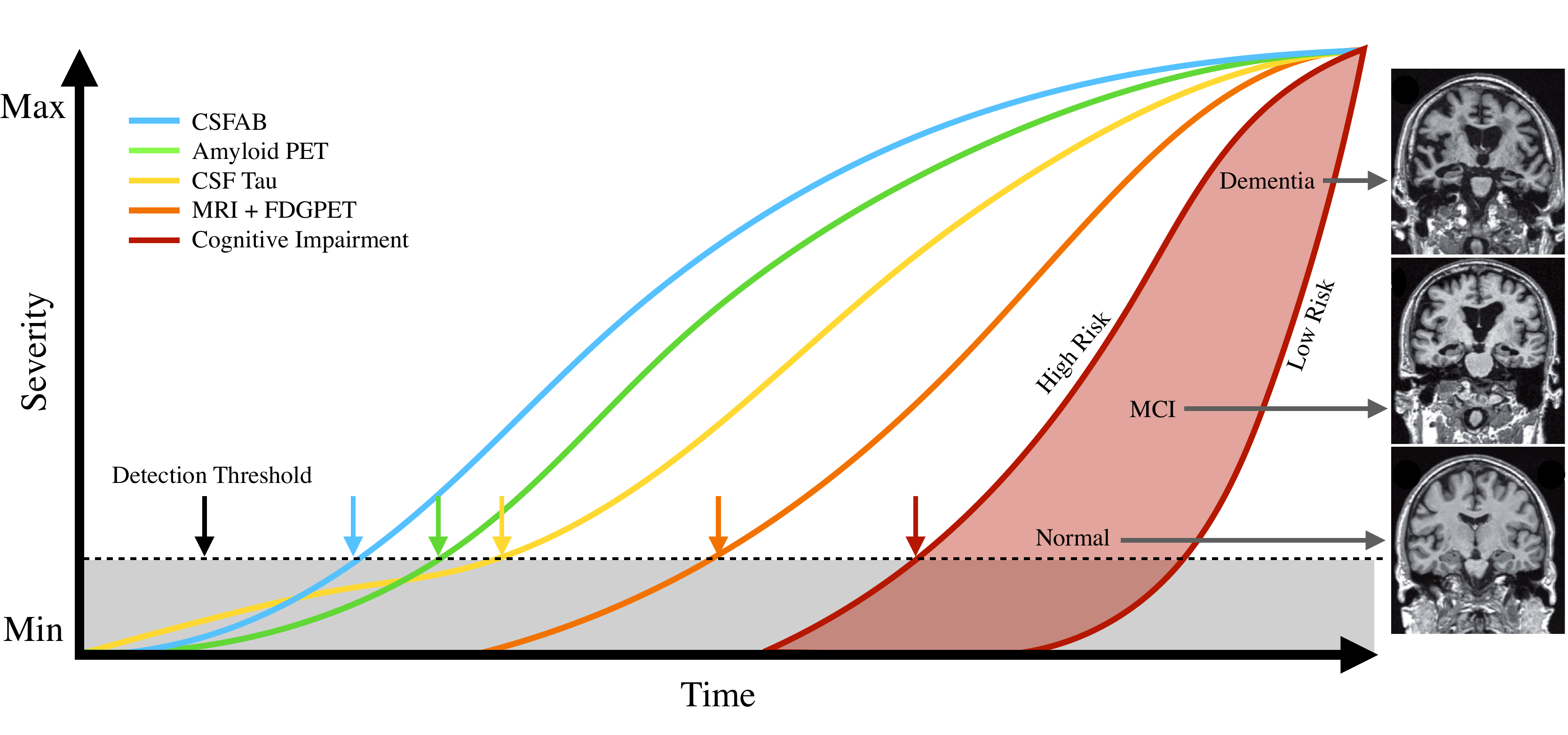}

\caption{\textbf{Stages of Alzheimer's Disease:} 
(left) Alzheimer's disease (AD) is a multifaceted disease in which cumulative pathological brain insults result in progressive cognitive decline that ultimately leads to dementia. Amyloid plaques, neurofibrillary tangles (NFTs), neurodegeneration, and inflammation are the well-established pathological hallmarks of AD. (right) Progressive atrophy (medial temporal lobes) in an older cognitively normal (CN) subject, an amnestic mild cognitive impairment (aMCI) subject, and an Alzheimer’s disease AD subject. \cite{khanahmadi2015genetic, Vemuri2010RoleOS}
}
\end{center}
\label{fig:ad-stages}
 \vspace{-2em}
\end{figure*}

%% file: sec/2_model.tex

\section{Forward Model of AD}
\label{sec:model}

\subsection{Problem setup}
In clinical trials, multiple cognitive tests are performed to assess current patient function and likelihood of AD progression in terms of different endpoints including CDRSB, MMSE, and ADAS-COG12. Our goal is to predict the progression of Alzheimer’s Disease 12 months from the first patient visit (baseline visit). Specifically, our model takes in as input a single 3D MR image (H x W x T x C) and D length vector containing the concatenation of cognitive scores, genomics, and demographic information at the baseline visit to predict the interpolated CDRSB,  MMSE and ADAS-COG12 at month 12 (i.e. after one year). The interpolated cognitive scores are computed by modeling the change in each cognitive score over time with a per-patient linear regression on all visits in the first 24 months with the baseline visit assigned a value of 0. We arrive at the interpolated value for a cognitive score by using the linear regression model’s prediction at 12 months. Training our models using the interpolated cognitive scores helps to mitigate missing values and measurement noise.

\subsection{Proposed deep learning pipeline}
We propose a multimodal multi-task deep neural network that combines 3D MR images and clinical features including cognitive scores, genomics,  and demographic information into a single end-to-end model to predict the future status of an AD patient from their baseline visit. Each modality is fed into the network through separate branches to learn modality-specific latent spaces, while leveraging complementary knowledge between different modalities. 3D MR images are fed into a 3D dense convolutional neural network that was pretrained on longitudinal data (respecting patient splitting) to predict current cognitive scores. A regularization loss based on mutual information \cite{kim2019learning}, was adopted to minimize the undesirable effects of the possible study-specific biases. Moreover, sharpness-aware minimization\cite{foret2020sharpness} was employed to improve model generalization and reduce the risk of overfitting. Finally, through an adversarial training process \cite{ganin2015unsupervised}, the model learns how to predict endpoints independent of domain shifts.

\subsection{Network Design}
The proposed model contains two main networks as seen in Figure 2.  The first is an encoder a, which learns an effective representation of the clinical information. The second is a combined network that consists of three subnetworks: feature extraction network f, endpoint prediction network g, and domain adaptation network h. The output of the feature extraction network enters both the endpoint prediction and domain adaptation networks. The three subnetworks are trained in an end-to-end manner, where the domain adaptation network h learns to remove the bias from embedded features learned by the imaging feature extraction network f to improve the endpoints predicted by $g \circ f$ \cite{kim2019learning,ganin2016domain}. 

The feature extraction network is a compact version of Densenet121\cite{huang2017densely}, designed by stacking four layers including a 3D convolutional layer, a 3D batch normalization layer, a leaky ReLU layer, and a 3D max-pooling layer followed by two dense blocks respectively with 6 and 12 dense layers. The dense layer is composed of 6 layers including a 3D batch normalization layer, a leaky ReLU layer, and a 3D convolution layer, where this set of 3 layers is repeated twice. The feature map size changes between the dense blocks through transition layers. The transition layer is composed of a 3D batch normalization layer, a leaky ReLU layer, a 3D convolution layer, and a 3D average pooling layer.
The endpoint prediction network $g$  is a single dense block with 16 dense layers. This network is trained to predict the endpoints, while encouraging $h$ to output a uniform prediction for all imaging sites. The adversarial network $h$ is also a single dense block with 16 dense layers, a leaky ReLU layer, a 3D average pooling layer, and a linear layer. This branch serves as a domain adaptation network, enabling the model to learn domain invariant imaging feature representations.

\input{fig/deepAD}

\subsection{Training}
The input patient $p$ with clinical features $Clin_{p}$ and a 3D magnetic resonance imaging $MRI_{p}$ is fed into the deep model, where $Clin_{p}$ enters the encoder $a$ and $MRI_{p}$ enters the dense network $f$ to respectively learn clinical embeddings $a(Clin_{p})$ and image embeddings $f(MRI_{p})$ of the patient. Subsequently, $f(MRI_{p})$ is fed into the domain adaptation network $h$ while both extracted features, $a(Clin_{p})$ and $f(MRI_{p})$, are fed forward through the endpoint prediction network $g$. The parameters of each network are defined as $\theta_{a}$, $\theta_{f}$, $\theta_{g}$, $\theta_{h}$ with the subscripts indicating their specific network. 

The objective of our model is to train a network that performs robustly on test data from an unseen domain, even though the network is trained with a mixture of data sources. To this end, we add mutual information based loss to the objective function for training networks as proposed in \cite{kim2019learning}. Therefore, the training procedure is to optimize the following problem:

\begin{align}
\min _{\theta_{a}, \theta_{f}, \theta_{g}}&\sum_{p \varepsilon P} \operatorname{loss}\left(y_{\text {endpoint }}(p), g\left(f\left(MRI_{p}\right) +a\left(Clin_{p}\right)\right)\right)\nonumber\\
&+\lambda \mathrm{I}\left(y_{\text {bias }}(p) ; f\left(MRI_{p}\right)\right)
\end{align}


where $loss(.,.)$ and $I(.,.)$ respectively indicate the loss function and the mutual information loss, and $\lambda$ is a hyper-parameter to balance the terms. Replacing the mutual information, the above equation becomes the following equation:

\begin{align}
\label{eqn:objective}
\min _{\theta_{a}, \theta_{f}, \theta_{g}} \max _{\theta_{h}} &\sum_{p}\left[\mathcal{L}_{\text {endpoint }}(p)-\mu \mathcal{L}_{\text {bias }}(p)\right. \nonumber \\
&\mathrel{\phantom{=}} \left.\kern-\nulldelimiterspace +\lambda H\left(h \circ f\left(M R I_{p}\right)\right)\right] \nonumber \\
\mathcal{L}_{\text {endpoint }}(p)&=\operatorname{loss}\left(y_{\text {endpoint }}(p), g \left(f\left(M R I_{p}\right)+a\left( \operatorname{Clin}_{p}\right)\right)\right) \nonumber \\
\mathcal{L}_{\text {bias }}(p)&=\operatorname{loss}\left(y_{\text {bias }}(p), h \circ f\left(M R I_{p}\right)\right) \nonumber \\
H(h \circ f(x))&=\sum_{x} h \circ f(x) \times \log (h \circ f(x))
\end{align}



where $\mathcal{L}_{\text {endpoint }}(.)$,  $\mathcal{L}_{\text {bias }}(.)$, and $H(h \circ f(.))$ respectively represent the loss of the endpoint prediction, the loss of the bias prediction, and the entropy of the bias which acts as a regularizer.  The set of networks, $a$, $f$, $g$, and $h$ are trained end to end, with the adversarial strategy \cite{goodfellow2014generative,chen2016infogan} and gradient reversal technique \cite{ganin2016domain} updating the image only network weights $\theta_{f}$, $\theta_{h}$. Early in learning, $g \circ f$ is rapidly trained to predict the endpoints by using the bias information. Then $h$ learns to predict the bias, and $f$ begin to learn how to extract feature embedding invariant to the imaging domain. 

\subsection{Inter-study biases}
There are two common forms of domain shifts (biases) in medical image analysis among different studies: population shift and acquisition shift. The population shift occurs when cohorts of subjects exhibit varying demographic or clinical characteristics, while the acquisition shift is observed due to differences in imaging protocols, modalities or scanners. 
To alleviate possible inter-study biases, our model regulates an additional network to minimize the mutual information shared between the extracted feature and inter-study distribution shift we want to unlearn (see equation \ref{eqn:objective}). The additional network is trained adversarially against the feature embedding network and predicts the bias distribution. At the end of learning, the domain prediction network is not able to predict the domain the image came from not because it is poorly trained, but because the feature embedding network successfully unlearns the bias information.


\subsection{Handling of missing data}
Patients without valid MR imaging and segmentations were excluded from the dataset for all experiments. Missing entries for continuous valued clinical inputs were imputed using a multivariate iterative imputation method\cite{zhang2016multiple} learned on the training data split and applied to the remaining splits. Missing categorical inputs were one-hot encoded as an additional class. As described in the problem setup, interpolation of labels using all timepoints between baseline to 24 months reduced the number of missing labels. Patients with no valid label (single task) or labels (multi task) were excluded, while partially missing labels were handled through masking of the loss function.

%% file: fig/deepAD.tex
\begin{figure*}
\begin{center}
\includegraphics[width=1.99\columnwidth]{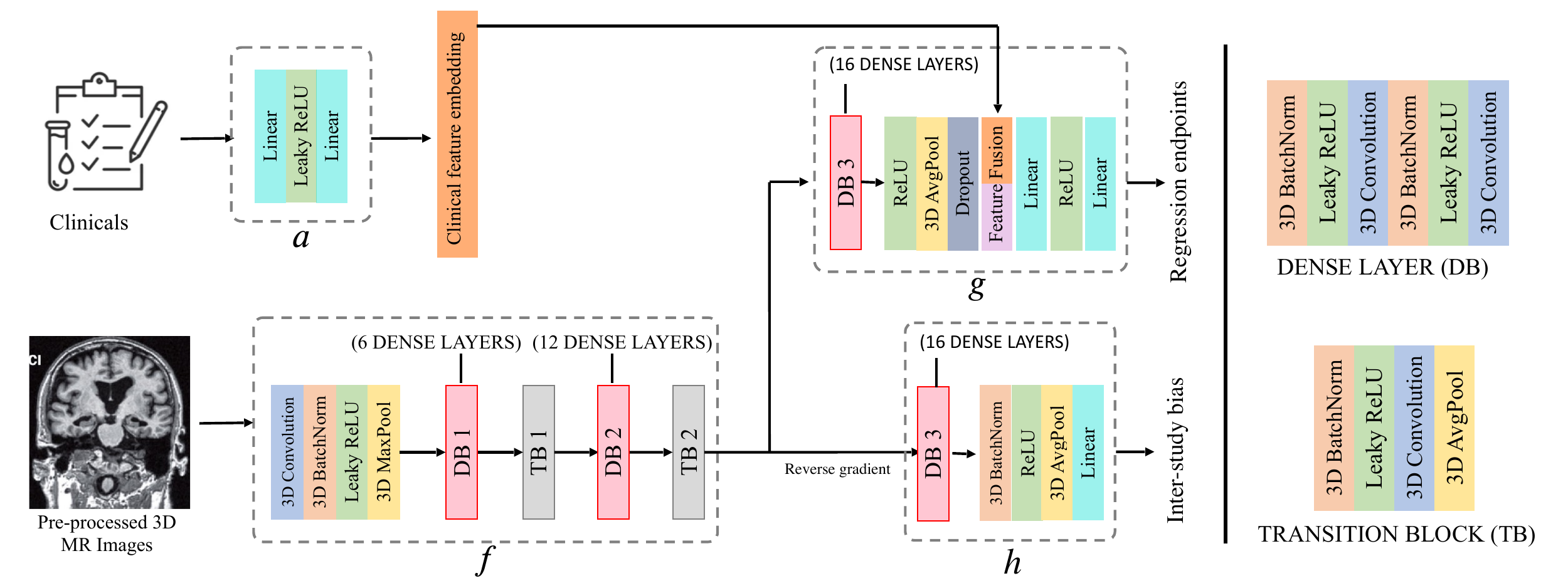}
\end{center}
\caption{
\textbf{DeepAD Architecture:} Illustration of the DeepAD model for predicting Alzheimer’s disease progress prediction. (left) DeepAD takes as input both clinical information (including cognitive scores, demographics, genomics) and 3D MR images and learns to predict the progression of the disease after 12 months (right) The dense layer is composed of 6 layers including a 3D batch normalization layer, a leaky ReLU layer, and a 3D convolution layer. The transition layer is composed of a 3D batch normalization layer, a leaky ReLU layer, a 3D convolution layer, and a 3D average pooling layer. Abbreviations: DB, Dense Block; TB, Transition Block}
\label{fig:DeepAD}
 \vspace{-1em}
\end{figure*}

%% file: sec/3_data.tex
\section{Datasets}
\label{sec:datasets}

All individuals included in our analysis were 5548 participants with 34731 total visits from the Alzheimer’s Disease Neuroimaging Initiative (ADNI) and six internal studies including A,B,C,D,E and F (see Table 1, dataset names are anonymized in order to allow for blind review). All studies were filtered to include patients who are positive Amyloid Beta, have a MMSE larger than 20 and are diagnosed as prodromal or mild at baseline. Amyloid positivity is detected by either cerebrospinal fluid (SCF) or positron emission tomography (PET). ADNI was launched in 2003 as a public–private partnership and their primary goal has been to test whether MR imaging, PET, other biological markers and clinical, and neuropsychological assessment can be combined to measure the progression of AD. For up-to-date information, see http://adni.loni.usc.edu. 


For each patient we have the following modalities:
\input{tab/datasets}

\begin{itemize}[noitemsep]
\item \textbf{Demographic:} Age, Sex, Diagnosis, Education level, and BBMI
\item \textbf{Genomic:} APOE4
\item \textbf{Cognitive scores:} CDRSB, MMSE, ADAS-COG12, and FAQ 
\item \textbf{Imaging:} Raw 3D MRI images
\end{itemize}

MR volumes were standardized using the following preprocessing steps. First, a brain mask was inferred for each volume using SynthSeg (https://github.com/BBillot/SynthSeg), a deep learning segmentation package. During training, the volumes and segmentations were resampled isotropically to 1mm voxel size, standardized to canonical (RAS+) orientation, intensity rescaled to 0,1 and Z-score normalized using only voxels inside the brain mask. Finally, volumes are cropped or padded to (150,185,155) using the TorchIO library (https://github.com/fepegar/torchio).

%% file: tab/datasets.tex
\begin{table}
\centering
\resizebox{7 cm}{!}{ 
\begin{tabular}{@{}lcccc@{}}
\toprule
Study & Abbreviation & \multicolumn{3}{c}{Number of patients} \\
 - & - & CN & MCI & AD \\
\midrule
$\text{Anon}_A$ & A & - & - & 374 \\
$\text{Anon}_B$ & B & - & - & 66 \\
$\text{Anon}_C$ & C & - & - & 197 \\
$\text{Anon}_D$ & D & - & 702 & - \\
$\text{Anon}_E$ & E & - & 240 & 437 \\
$\text{Anon}_F$ & F  & - & 271 & 341 \\
ADNI & ADNI & 272 & 490 & 185  \\
\bottomrule
\end{tabular}
}
\caption{\textbf{Summary of datasets:} Abbreviations: CN, cognitively normal; MCI, mild cognitive impairment; AD Alzheimer’s disease.}
\label{tab:datasets}
\vspace{-1em}
\end{table}



%% file: sec/4_results.tex
\section{Results}
\label{sec:results}

The training data includes all studies except E and F. It is then split into a training set (50\%) and a validation set (50\%). Two different test sets were prepared, an in-study test set and out-study test set where the first one includes unseen patients from the same studies in training data and the latter includes patients from E and F studies that are not used for training and validation.

Different setups were tested to evaluate the effectiveness and performance of our proposed multimodal multi-task deep learning model. More experiments were conducted to explore the importance of adversarial loss, and sharpness-aware minimization. To measure the performance of our model in different setups, a weighted coefficient of determination is defined separately for each endpoint by the weighted average of the coefficient of determinations on each dataset:

\begin{align}
R^{2}=\frac{\sum_{\substack{\text {d in dataset }}} R_{d}^{2} \times|d|}{\sum_{\substack{\text {d in dataset }}}|d|}
\end{align}

where $R^{2}_{d}$ indicates the coefficient of determination for dataset $d$.  This weighting guarantees that $R^{2}_{d}$ contributes to the weighted $R^{2}$ in proportion to the dataset size. Pooling all datasets and determining an aggregate $R^{2}$ is not desired given the artificially high performance resulting from different disease states within each dataset.

\input{tab/detail}

\subsection{Single modality single task modeling (SMST)}
Before combining all modalities, the performance of a single modality for predicting each endpoint was evaluated on the validation set. The results are also compared with an expert curated linear regression model. As can be noticed in Table ~\ref{tab:detail}, the best performance in terms of $R^{2}$ for interpolated CDRSB, MMSE, and ADAS-COG12 are obtained by DeepAD applied on clinical information, indicating that  baseline cognitive scores, genetics, and demographics are significant contributors to predicting progression.  
To mitigate the problem of having a small training set, we take advantage of longitudinal data and pretrain DeepAD to predict current cognitive scores for each visit of the patient. The pretrained weights are loaded into the imaging backbone then fine-tuned to predict future cognitive scores. As expected, the pretrained model yields better performance compared to the randomly initialized models.   

\subsection{Single modality multitask modeling (SMMT)}
The main caveat of single task modeling is that cognitive scores are predicted separately without considering their inherent correlations which could boost individual task performance. Besides, it is not clinically practical to develop a separate model for each endpoint, specifically when there are many endpoints to be predicted. 
In this section, the performance of each modality for multitask prediction of interpolated cognitive scores is reported on the validation set. The model incorporating only clinical information yields a smaller $R^{2}$ for interpolated ADAS-COG12 and interpolated MMSE and does not show an improvement for interpolated CDRSB compared to the values in SMST. This might suggest that the way that clinical information used to make predictions is different for each task.  On the other hand, the performance of DeepAD incorporating only MR images, either trained from random initialization or pretrained, is superior to that in SMST setup for each individual interpolated cognitive score, suggesting that related cognitive scores share a common relevant structural feature subset in MR images. DeepAD in SMMT setup yields up to 18\% and 13\% improvement in $R^{2}$ for interpolated MMSE and ADAS-COG12 over DeepAD in SMST setup.

\subsection{Multimodal single task modeling (MMST)}
To evaluate the importance of multi modal data, the performance of our model was assessed when predicting each endpoint separately. The model incorporating both clinical information and 3D MRIs yields a $R^{2}$ equal to 0.26, 0.15 and 0.20 for interpolated CDRSB, MMSE, and ADAS-COG12 respectively which is significantly larger than those in all experiments in both SMST and  SMMT. These results indicate that DeepAD is able to effectively utilize the correlation among the  modalities to considerably improve the prediction of the disease progression.

\subsection{Multimodal multitask modeling (MMMT)}
In this experiment, the adversarial loss and the sharpness-aware minimization are evaluated when both modalities are fed to the model to predict all three cognitive scores. The results in Table ~\ref{tab:detail} shows that DeepAD leads to the best performance when it utilizes all modalities to predict all three endpoints.

\subsection{Raw MRI vs hard-coded features}
\begin{figure}
\includegraphics[width=\linewidth]{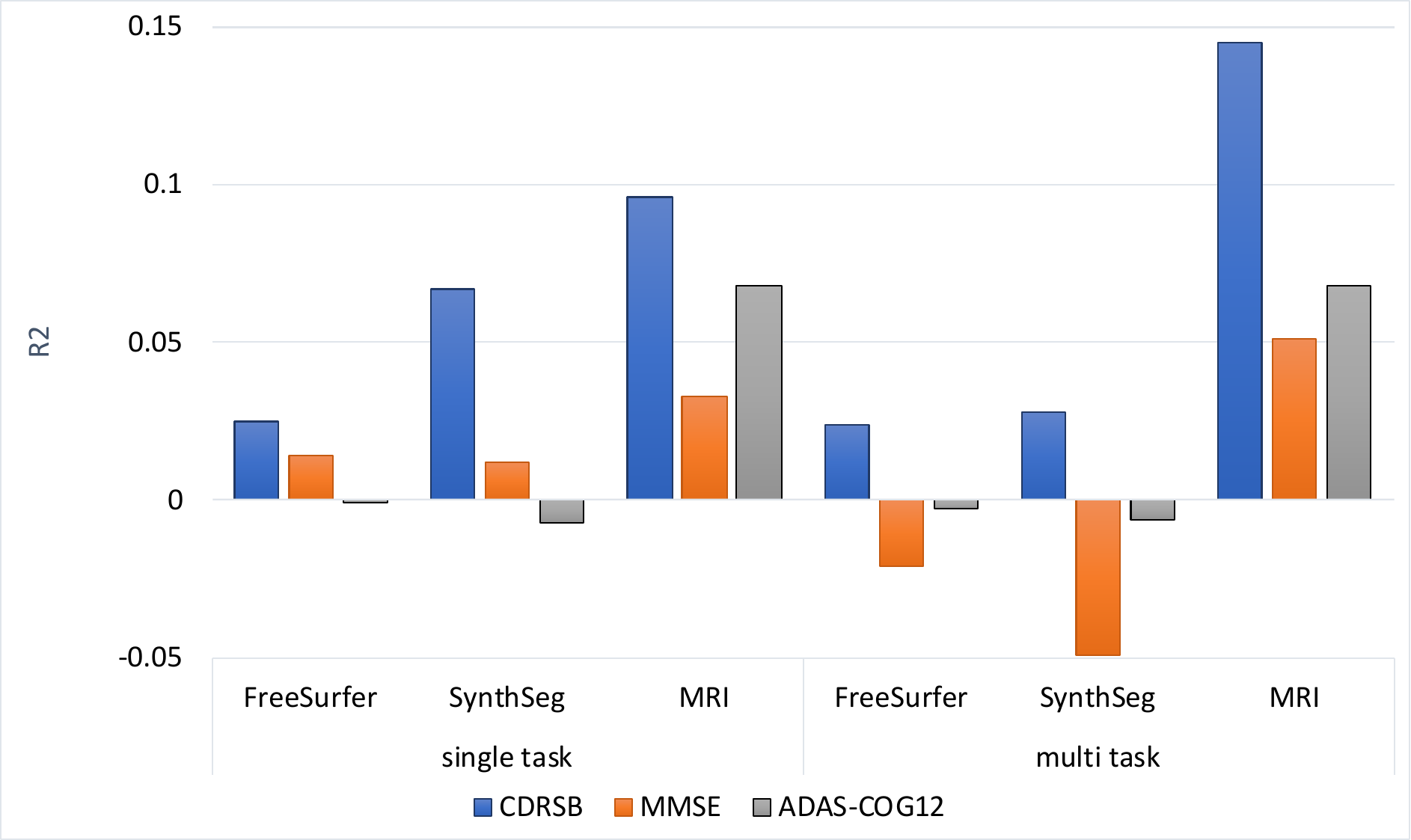}
\caption{Performance of DeepAD applied on extracted volumetric features (extracted by FreeSurfer or SynthSeg) or MRIs in terms of $R^{2}$ in both single task and multi task setups }\label{fig:vol}
\vspace{-1em}
\end{figure}

To further understand the potential of MR images in predicting Alzheimer's prediction, several  single modality single task modeling experiments are conducted by using volumetric features including brain volume, hippocampus volume, and ventricle volume that are extracted either by FreeSurfer v6.0 \cite{fischl2012freesurfer,palumbo2019evaluation} or SynthSeg \cite{billot2021synthseg}. The results in Figure ~\ref{fig:vol} show that extracted volumetric features do not provide compelling accuracy in predicting cognitive endpoints neither in single task nor in multitask setups, indicating that they do not take advantage of correlation among cognitive scores. Figure  ~\ref{fig:vol} suggests that raw images encode structural features beyond what is captured by volumetric features for predicting the progression of the disease.

\subsection{In- and out-of-domain generalization performance}

The robustness of DeepAD is evaluated by using two different external test sets including in-study test set and out-study test set. The trained multi-task DeepAD was tested by using either clinical information or MRI or both. As presented in Table ~\ref{tab:test}, DeepAD applied on both modalities outperforms the other models significantly on in-study test-sets. On out-study test set DeepAD applied on both modalities significantly outperforms the other models for predicting CDRSB,  while reaching a comparable performance for MMSE and ADAS-COG12 prediction when applied only on clinical information.        

Moreover, stratifying patient cognitive trajectories by their predicted CDSB change (see Supplementary Material) reveals marked differences between groups as DeepAD could separate stable from declining patients.

\input{tab/test_set}

\subsection{Other results}
\begin{figure}
\includegraphics[width=\linewidth]{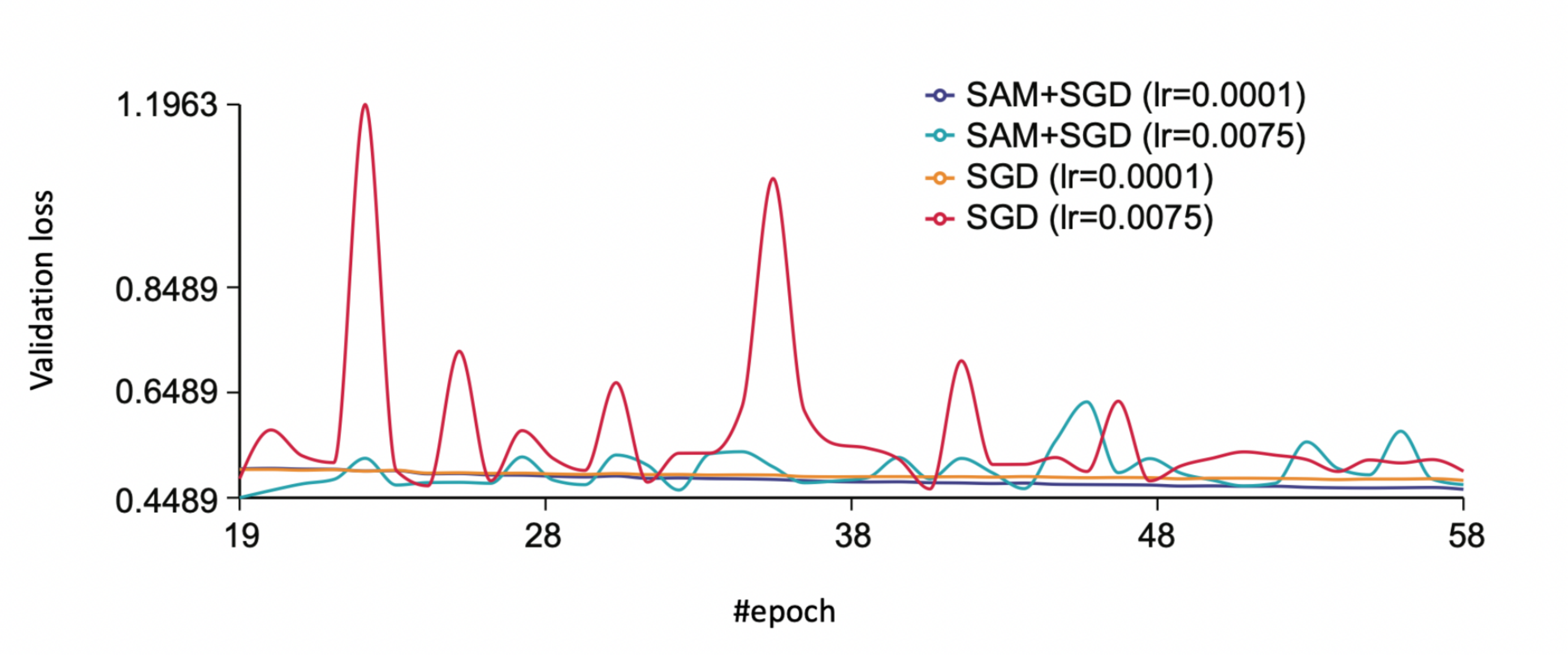}
\caption{Robustness of different optimization methods: SAM on top of SGD leads to a more stable convergence and better minima.}\label{fig:sam}
\vspace{-1em}
\end{figure}

In this section, the sharpness-aware minimization and adversarial loss are evaluated when both modalities are fed to the model to predict all three cognitive scores. 
Figure ~\ref{fig:sam} shows the validation loss by optimizing with either SGD or SAM+SGD using two different learning rates. We magnified part of the training procedure to provide more visibility. It is observed that training with SGD is very unstable near convergence and thus prevents the model from converging correctly.  SGD and SAM+SGD with a learning rate equal to 0.001 have the stable behavior with a minimum loss equal to 0.5 and 0.4 respectively which yields to $R^{2}$ equal to 0.16 and 0.27 respectively.  

To explore the effectiveness of adversarial loss for promoting robustness, DeepAD is tested on datasets E and F in a MMMT setup with and without incorporating adversarial loss. As presented in Table  ~\ref{tab:robust}, DeepAD incorporating adversarial loss yield up to 0.22\% and 0.09\% in terms of $R^{2}$ for MMSE prediction on studies E and F which are not included for training and validation.  

\input{tab/robust}

%% file: tab/detail.tex
\begin{table}[ht]
\centering
\resizebox{\linewidth}{!}{ 
\begin{tabular}{@{}l|l|ccc@{}}
\toprule
& Method-modality & CDRSB & MMSE & ADAS-COG12 \\
\midrule
\multirow{4}{*}{SMST} & DeepAD-Clin & 0.20 & 0.11 & 0.12 \\
                      & DeepAD-MRI - rand init & 0.10 & 0.03 & 0.07 \\
                      & DeepAD-MRI - pretrained & 0.15 & 0.09 & 0.13 \\
                      & Regression-Clin & 0.17 & 0.04 & 0.08 \\
\midrule
\multirow{3}{*}{SMMT} & DeepAD-Clin & 0.20 & 0.10 & 0.06 \\
                      & DeepAD-MRI - rand init & 0.14 & 0.05 & 0.07 \\
                      & DeepAD-MRI - pretrained & 0.16 & 0.11 & 0.15 \\
\midrule
\multirow{1}{*}{MMST} & DeepAD-MRI+Clin & 0.26 & 0.15 & 0.20 \\
\midrule
\multirow{1}{*}{MMMT} 
                      & DeepAD-MRI+Clin & 0.26 & 0.21 & 0.21 \\
\bottomrule
\end{tabular}
} 
\caption{
\textbf{Detailed validation results:}
Performance of DeepAD on validation set across inputs and tasks. Stability of performance across 5 random seeds is reported in the supplementary material. Abbreviation: SMST, single modality single task modeling separately for each interpolated cognitive score in terms of $R^{2}$; SMMT, single modality multi task modeling for jointly predicting three interpolated cognitive scores in terms of $R^{2}$; MMST, multi modality single task modeling; MMMT, multi modality multi task modeling; Clin: Clinical information modality; MRI, imaging modality  
} 
\label{tab:detail}
\vspace{-1em}
\end{table}

%% file: tab/test_set.tex
\begin{table}[h]
\centering
\resizebox{\linewidth}{!}{ 
\begin{tabular}{@{}l|l|cccc@{}}
\toprule
endpoint & study & Linear regression & DeepAD-Clin & DeepAD-MRI & DeepAD-MRI+Clin \\
\midrule
\multirow{2}{*}{CDRSB} & in-study & 0.17 & 0.15 & 0.12 & \textbf{0.22} \\
                      & out-study & 0.12 & 0.11 & 0.08 & \textbf{0.17} \\
\midrule
\multirow{2}{*}{MMSE} & in-study & 0.09 & 0.09 & 0.14 & \textbf{0.18} \\
                      & out-study & \textbf{0.13} & 0.10 & 0.03 & 0.10\\
\midrule
\multirow{2}{*}{ADAS-COG12} & in-study & 0.06 & 0.06 & 0.11 & \textbf{0.15} \\
                      & out-study & 0.06 & 0.07 & \textbf{0.08} & \textbf{0.08} \\
\bottomrule
\end{tabular}
} 
\caption{
\textbf{Generalization performance:}
DeepAD exhibits strong robustness to domain shift in the out of study test set. Clin: Clinical information modality; MRI, imaging modality  
} 
\label{tab:test}
\vspace{-1em}
\end{table}

%% file: tab/robust.tex
\begin{table}
\centering
\resizebox{\linewidth}{!}{ 
\begin{tabular}{@{}lcccccc@{}}
\toprule
endpoint & \multicolumn{2}{c}{study E} & \multicolumn{2}{c}{study F} &\multicolumn{2}{c}{\% of improvement} \\
  & with & without & with & without & study E & study F\\
\midrule
CDRSB & 0.18 & 0.17 & 0.16 & 0.15 & 0.02 & 0.03 \\
MMSE & 0.09 & 0.07 & 0.11 & 0.10 & 0.22 & 0.09\\
ADAS-COG12 & 0.11 & 0.11 & 0.05 & 0.05 & 0.00 & 0.00\\

\bottomrule
\end{tabular}
} 
\caption{
\textbf{Adversarial loss effect for improving robustness:} performance of deepAD with and without incorporating adversarial training as well as the percentage of improvement in terms of $R^{2}$ for studies E and F that are not used for training and validation.
} 
\label{tab:robust}
\vspace{-1em}
\end{table}

%% file: sec/5_conclusions.tex
\section{Conclusions}
\label{sec:conclusions}
We propose \textbf{DeepAD}, a multimodal multi-task deep learning approach to predict the progression of Alzheimer's Disease in terms of different endpoints by using longitudinal clinical features and raw neuroimaging data. We apply a 3D convolutional neural network to extract the spatiotemporal features of MR images and then integrate those features with other information sources. In order to alleviate inter-study domain shift and improve generalization, DeepAD utilizes an adversarial loss and sharpness-aware-minimization. Our result show an improvement in prediction accuracy and more robust prediction performance for patients in early stages of Alzheimer's disease. The proposed multimodal multi-task deep learning approach has potential to identify patients at higher risk of progressing to AD and help develop better therapies at lower cost to society.

\section{Acknowledgements}
We would like to thank all of the study participants and their families, and all of the site investigators, study coordinators, and staff. Assistance in preparing this article for publication was provided by Genentech, Inc. 

Part of data collection and sharing for this project was funded by the ADNI (National Institutes of Health Grant U01 AG024904). ADNI is funded by the National Institute on Aging (NIA), the National Institute of Biomedical Imaging and Bioengineering (NIBIB), and through generous contributions from the following: Alzheimer's Association; Alzheimer's Drug Discovery Foundation; BioClinica; Biogen Idec; Bristol-Myers Squibb Company; Eisai; Elan Pharmaceuticals; Eli Lilly and Company; F. Hoffmann-La Roche and its affiliated company Genentech, Inc.; GE Healthcare; Innogenetics NV; IXICO; Janssen Alzheimer Immunotherapy Research \& Development, LLC.; Johnson \& Johnson Pharmaceutical Research \& Development LLC.; Medpace; Merck \& Co.; Meso Scale Diagnostics, LLC.; NeuroRx Research; Novartis Pharmaceuticals Corporation; Pfizer; Piramal Imaging; Servier; Synarc; and Takeda Pharmaceutical Company. The Canadian Institutes of Health Research is providing funds to support ADNI clinical sites in Canada. Private sector contributions are facilitated by the Foundation for the National Institutes of Health (www.fnih.org). The grantee organization is the Northern California Institute for Research and Education, and the study is coordinated by the Alzheimer's Disease Cooperative Study at the University of California, San Diego. ADNI data are disseminated by the Laboratory for NeuroImaging at the University of California, Los Angeles

%% file: sec/X_supplementary.tex
\appendix




\section{Additional experiments}
\subsection{Stratification of predicted cognitive trajectories}
\begin{figure}
\includegraphics[width=\linewidth]{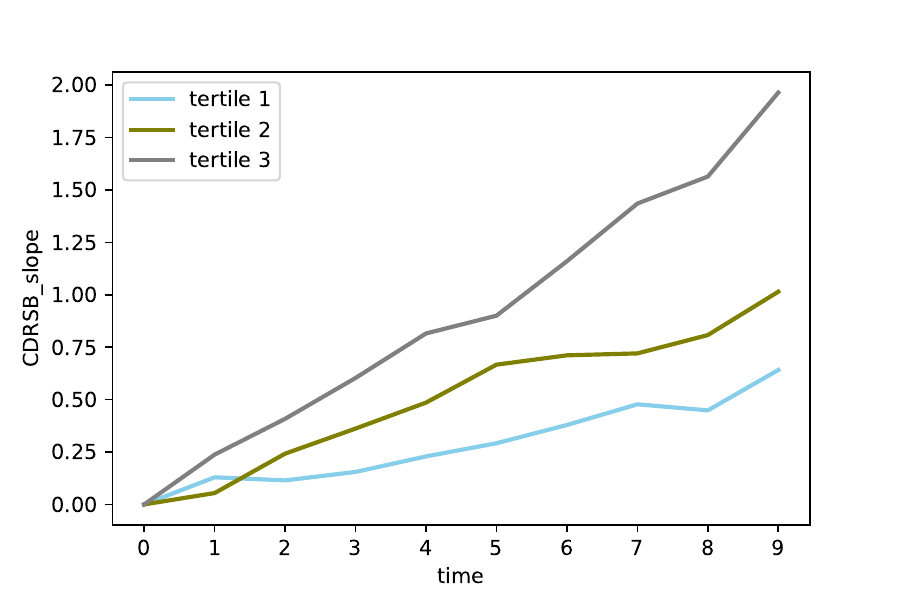}
\caption{Stratification of predicted cognitive trajectories when DeepAD(Clin+MRI) is used for inference. Plot illustrated the mean change in cognitive function measured by interpolated CDRSB over time (months*3).}\label{fig:kp-cdsb-CLIN-MRI}
\vspace{-1em}
\end{figure}

\begin{figure}
\includegraphics[width=\linewidth]{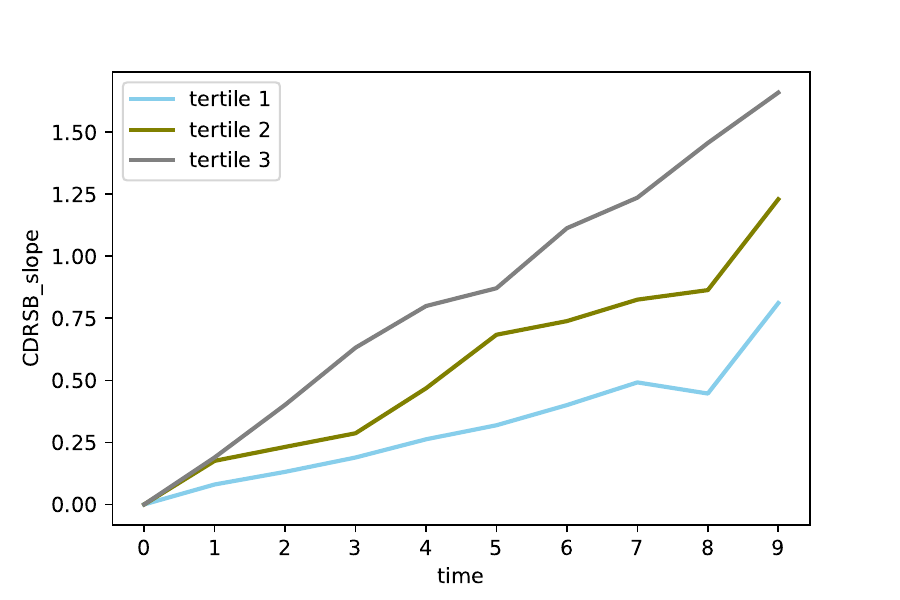}
\caption{Stratification of predicted cognitive trajectories when DeepAD(MRI) is used for inference. Plot illustrates the mean change in cognitive function measured by observed(actual) interpolated CDRSB over time (months*3).}\label{fig:kp-cdsb-MRI}
\end{figure}

The patients in the out-study test set were divided into tertiles based on predicted interpolated CDRSB. Tertile 3 (greatest predicted interpolated CDRSB) was associated with faster disease progression compared to other tertiles. Figures ~\ref{fig:kp-cdsb-CLIN-MRI} and ~\ref{fig:kp-cdsb-MRI} respectively present the plots for the out-study test when DeepAD(Clin+MRI) and DeepAD(MRI) models are used for inference. Stratified analysis demonstrates DeepAD predictions for interpolated CDRSB at 12 months (time=4) separates patients with different rates of progression up to 24 months (time=8).

\subsection{Performance of linear regression in multi modal single task setup}
The results of linear regression model applied on both clinical information and volumetric features (extracted by FreeSurfer) are presented in Table ~\ref{tab:lin}. Linear regression models are not able to use 3D MR images as input and yet by incorporating volumetric features can not outperform DeepAD.              

\begin{table}
\centering
\resizebox{\linewidth}{!}{ 
\begin{tabular}{@{}l|l|ccc@{}}
\toprule
& Method-modality & CDRSB & MMSE & ADAS-COG12 \\
\midrule
\multirow{1}{*}{MMST} & Regression-Clin+Volumetric & 0.18 & 0.17 & 0.09 \\
\bottomrule
\end{tabular}
} 
\caption{Performance of linear regression in multi modal single task setup}
\label{tab:lin}
\end{table}

\section{Stability across random seeds}
Performance of DeepAD on the validation set across inputs and tasks for 5 random seeds, reported as mean (standard deviation) weighted $R^2$ for interpolated CDRSB, interpolated MMSE, and interpolated ADAS-COG12.
\vspace{1em}


\centering
\resizebox{\linewidth}{!}{ 
\begin{tabular}{@{}l|l|ccc@{}}
\toprule
& Method-modality & CDRSB & MMSE & ADAS-COG12 \\
\midrule
\multirow{2}{*}{SMST} & DeepAD-MRI - rand init & 0.0345 (0.01960) & 0.0395 (0.02410) & 0.0452 (0.00852) \\
                      & DeepAD-MRI - pretrained & 0.1550 (0.00301) & 0.1660 (0.00879) & 0.1300 (0.00206) \\
                      
\midrule
\multirow{2}{*}{SMMT} & DeepAD-MRI - rand init & 0.1310 (0.00876) & 0.1090 (0.02620) & 0.1080 (0.02900) \\
                      & DeepAD-MRI - pretrained & 0.1610 (0.00315) & 0.1190 (0.01030) & 0.1550 (0.00844) \\
\midrule
\multirow{1}{*}{MMST} & DeepAD-MRI+Clin & 0.2600 (0.00629) & 0.2280 (0.00522) & 0.1940 (0.01320) \\
\midrule
\multirow{1}{*}{MMMT} & DeepAD-MRI+Clin & 0.2470 (0.00756) & 0.2010	(0.01040) & 0.2170 (0.00740) \\

\bottomrule
\end{tabular}
} 

                      